\documentclass{article}

\PassOptionsToPackage{numbers, compress}{natbib}
\usepackage[preprint]{neurips_2025}

\usepackage[utf8]{inputenc} 
\usepackage[T1]{fontenc}    
\usepackage{hyperref}       
\usepackage{url}            
\usepackage{booktabs}       
\usepackage{amsfonts}       
\usepackage{nicefrac}       
\usepackage{microtype}      
\usepackage{xcolor}         
\usepackage{graphicx}
\usepackage{amsmath}
\usepackage{bbm}
\usepackage{caption}
\usepackage{booktabs}
\usepackage{multirow}

\title{UAV-ON: A Benchmark for Open-World \\ Object Goal Navigation with Aerial Agents}

\author{
	\textbf{
		Jianqiang Xiao\quad
		Yuexuan Sun\quad
		Yixin Shao\quad
		Boxi Gan} \\
	\textbf{
		Rongqiang Liu\quad
		Yanjin Wu\quad
		Weili Guan\quad
		Xiang Deng\thanks{Corresponding author}} \\
	Harbin Institute of Technology, Shenzhen \\
	\texttt{xiaojianqiang@hit.edu.cn}
}

\begin{document}

\maketitle

\begin{abstract}
Aerial navigation is a fundamental yet underexplored capability in embodied intelligence, enabling agents to operate in large-scale, unstructured environments where traditional navigation paradigms fall short. However, most existing research follows the Vision-and-Language Navigation (VLN) paradigm, which heavily depends on sequential linguistic instructions, limiting its scalability and autonomy. To address this gap, we introduce UAV-ON, a benchmark for large-scale Object Goal Navigation (ObjectNav) by aerial agents in open-world environments, where agents operate based on high-level semantic goals without relying on detailed instructional guidance as in VLN. UAV-ON comprises 14 high-fidelity Unreal Engine environments with diverse semantic regions and complex spatial layouts, covering urban, natural, and mixed-use settings. It defines 1270 annotated target objects, each characterized by an instance-level instruction that encodes category, physical footprint, and visual descriptors, allowing grounded reasoning. These instructions serve as semantic goals, introducing realistic ambiguity and complex reasoning challenges for aerial agents. To evaluate the benchmark, we implement several baseline methods, including Aerial ObjectNav Agent (AOA)—a modular policy that integrates instruction semantics with egocentric observations for long-horizon, goal-directed exploration. Empirical results show that all baselines struggle in this setting, highlighting the compounded challenges of aerial navigation and semantic goal grounding. UAV-ON aims to advance research on scalable UAV autonomy driven by semantic goal descriptions in complex real-world environments. Our benchmark and code are available at: \url{https://github.com/Kyaren/UAV_ON}.
\end{abstract}

\vspace{0.5em}

\section{Introduction}
In recent years, Unmanned Aerial Vehicles (UAVs) have been rapidly deployed across a wide range of applications, including cargo transportation ~\cite{golabi2022,barmpounakis2016}, emergency rescue operations ~\cite{zhao2024,scherer2015}, and environmental monitoring ~\cite{asadzadeh2022,liu2022}. With the rise of smart cities~\cite{mohamed2020} and advances in Low-Altitude Airspace Management (LAAM)~\cite{pongsakornsathien2025advances}, UAVs are increasingly expected to operate autonomously at scale in diverse real-world environments. To achieve this vision, UAVs are expected to go beyond basic flight control by developing the ability to perceive, understand, and navigate complex, dynamic, and unstructured environments ~\cite{wang2020}. Among these capabilities, intelligent navigation is particularly critical, as it determines whether a UAV can autonomously reach mission-critical targets under uncertainty and dynamic conditions~\cite{bijjahalli2020}.

Recent progress in aerial Vision-and-Language Navigation (VLN) \cite{liu2023aerialvln, chu2024, wang2024, gao2025, zhao2025} has been largely driven by detailed, fine-grained language instructions. However, such approaches often rely on externally provided step-by-step guidance, limiting their scalability in open-world environments where agents are expected to interpret high-level goals and navigate autonomously \cite{hong2025gsavln}. Object Goal Navigation (ObjectNav) presents a compelling alternative by requiring agents to locate target objects based on semantic cues, without depending on dense instruction sequences. While ObjectNav has been widely studied in ground-based indoor scenarios~\cite{kolve2017ai2thor, deitke2020robothor, ramakrishnan2021hm3d}, its potential in aerial navigation across large-scale, unstructured outdoor environments remains underexplored. Bridging this gap could open new avenues for deploying autonomous aerial systems capable of goal-driven exploration in complex real-world settings.

Accordingly, we introduce UAV ObjectNav (UAV-ON), a large-scale benchmark for evaluating autonomous, goal-driven navigation by UAVs in open-world environments. UAV-ON features a diverse range of outdoor settings, including urban landscapes, highways, forests, waterways, and mountainous terrain, designed to capture the semantic richness and visual complexity of real-world environments. Each task is specified through an instance-level semantic instruction that includes the target object's category, approximate size, and a descriptive natural language cue, enabling agents to infer and locate the appropriate object. In each episode, the UAV is randomly placed within the environment and required to navigate toward the target object using only egocentric RGB-D inputs from front, left, right, and downward-facing sensors. The agent performs obstacle avoidance and path planning entirely through onboard visual perception, without access to any global map or external information. UAV-ON serves as a standardized platform for studying autonomous exploration and object retrieval by UAVs in large-scale, complex environments.

To assess the UAV-ON benchmark, we implement a suite of baselines, including Aerial ObjectNav Agent (AOA)—a zero-shot modular policy built on a pre-trained Multimodal Large Language Model (MLLM). AOA unifies visual perception, semantic reasoning, obstacle avoidance, and action selection through a prompting-based control pipeline, allowing it to navigate complex environments without task-specific training. Our main contributions are summarized as follows:
\begin{itemize}
	\item We present UAV-ON, the first large-scale benchmark for instance-level ObjectNav by aerial agents in open-world environments. Unlike existing UAV-based VLN benchmarks that rely on dense step-by-step language supervision, UAV-ON defines over 11,000 navigation tasks using compact semantic goal instructions, spanning 14 high-fidelity outdoor scenes with diverse layouts and realistic object placements.
	\item UAV-ON requires aerial agents to execute discrete, parameterized actions through physically grounded simulation in cluttered outdoor environments, in contrast to the teleport-based control used in prior VLN benchmarks. Agents are expected to interpret high-level semantic goals and reason about object-scene co-occurrence, while simultaneously handling real-world challenges such as obstacle avoidance and collision dynamics.
	\item We introduce a set of baseline policies for UAV-based object-goal navigation, including a random policy, a CLIP-based semantic heuristic algorithm (CLIP-H), and our proposed Aerial ObjectNav Agent (AOA)—a zero-shot framework driven by a pre-trained multimodal language model. AOA encodes multimodal inputs into prompts and directly generates semantic action outputs with control magnitudes, featuring two variants: AOA (Fixed-step), which utilizes discrete, fixed-parameter actions, and AOA (Variable-step), allowing flexible control selection by the language model.
\end{itemize}

\section{Related Work}
\subsection{Object Goal Navigation}
ObjectNav tasks~\cite{anderson2018evaluation} require an agent to navigate towards a specified object category based solely on egocentric visual observations. Early benchmarks such as AI2-THOR ~\cite{kolve2017ai2thor} and Gibson ~\cite{xia2018gibson} provided the initial testbeds for studying embodied navigation in indoor environments. Later, RoboTHOR ~\cite{deitke2020robothor} and Habitat-Matterport 3D Dataset (HM3D) ~\cite{ramakrishnan2021hm3d} introduced more scalable and realistic simulation platforms, establishing standardized evaluation protocols for ObjectNav tasks. Methodologically, early works employed end-to-end reinforcement learning or imitation learning ~\cite{ye2021hiem, ye2021auxiliary, ramrakhya2023pirlnav}, followed by approaches incorporating semantic mapping and modular planning pipelines ~\cite{chaplot2020objectgoal,luo2022stubborn,campari2022abstractmodels}, and more recently, zero-shot ObjectNav leveraging Vision-Language Models (VLMs) to generalize towards unseen object categories ~\cite{zhao2023zeroshot, zhao2023semantic, chen2023zeroshot}. However, most prior efforts have focused on small-scale, static, and semantically homogeneous indoor environments with predominantly planar navigation. In contrast, we extend ObjectNav to large-scale, outdoor UAV scenarios, where agents operate in open-world landscapes, such as urban areas, forests, and water bodies, with three-dimensional movement and diverse semantic targets, presenting fundamentally new challenges in perception, planning, and generalization.
\subsection{Aerial Navigation}
Aerial navigation has attracted increasing interest in recent years, with early efforts centered on conventional tasks such as map-based planning and visual navigation~\cite{lu2018survey, aitsaadi2022uav, jones2023path}. More recently, the field has moved toward VLN, aiming to enable UAVs to understand and act on natural language instructions in large-scale, open-world environments. Notable contributions include AerialVLN~\cite{liu2023aerialvln}, the first UAV-centric VLN benchmark, as well as subsequent works like GeoText-1652~\cite{chu2024} and CityNav~\cite{lee2024citynav}, which focus on fine-grained geospatial grounding and the integration of geographic context. Simulation platforms such as TravelUAV~\cite{wang2024} and OpenFly~\cite{gao2025} have further advanced the field by providing scalable environments and extensive trajectory datasets, while methods like STMR~\cite{gao2024stmr}, GridVLN~\cite{zhao2025}, NavAgent~\cite{liu2024navagent}, and GeoNav~\cite{xu2025geonav} have introduced sophisticated representations for spatial reasoning. Existing aerial VLN approaches often rely on discrete waypoint execution or teleport-based movement, thus bypassing critical aspects of real-world navigation, such as obstacle avoidance, motion continuity, and collision handling~\cite{yue2024safe}. In contrast, we formulate a task where UAVs execute parameterized, physically grounded actions to explore open-world environments and locate semantic targets based on high-level instructions. This setup removes dependence on step-by-step guidance and better captures the physical and semantic complexities of real-world UAV operations, supporting more realistic evaluation of autonomous aerial systems.

\begin{table}[!t]
	\small
	\centering
	\caption{Comparison of ObjectNav and Aerial Navigation benchmarks.}
	\label{tab:dataset_comparison}
	\resizebox{0.8\linewidth}{!}{
		\begin{tabular}{ccccc}
			\toprule
			\textbf{Benchmark} & \textbf{Viewpoint} & \textbf{Task} & \textbf{Goal Type} & \textbf{Goal Specification} \\
			\midrule
			AI2-THOR~\cite{kolve2017ai2thor}          & Ground   & ObjNav   & Category          & Category Label \\
			Gibson~\cite{xia2018gibson}            & Ground   & ObjNav   & Category         & Category Label \\
			RoboTHOR~\cite{deitke2020robothor}          & Ground   & ObjNav   & Category          & Category Label \\
			HM3D~\cite{ramakrishnan2021hm3d}              & Ground   & ObjNav   & Category          & Category Label \\
			GeoText~\cite{chu2024}      & Aerial  & VLN      & Location          & Movement Instruction \\
			AerialVLN~\cite{liu2023aerialvln}         & Aerial  & VLN      & Location          & Movement Instruction \\
			CityNav~\cite{lee2024citynav}           & Aerial  & VLN      & Location          & Movement Instruction \\
			TravelUAV~\cite{wang2024}           & Aerial  & VLN      & Location          & Movement Instruction \\
			OpenFly~\cite{gao2025}           & Aerial  & VLN      & Location          & Movement Instruction \\
			\textbf{UAV-ON (Ours)} & \textbf{Aerial}  & \textbf{ObjNav}   & \textbf{Instance} & \textbf{Semantic Instruction} \\
			\bottomrule
		\end{tabular}
	}
\end{table}

\begin{figure}[htbp]
	\centering
	\includegraphics[width=\linewidth]{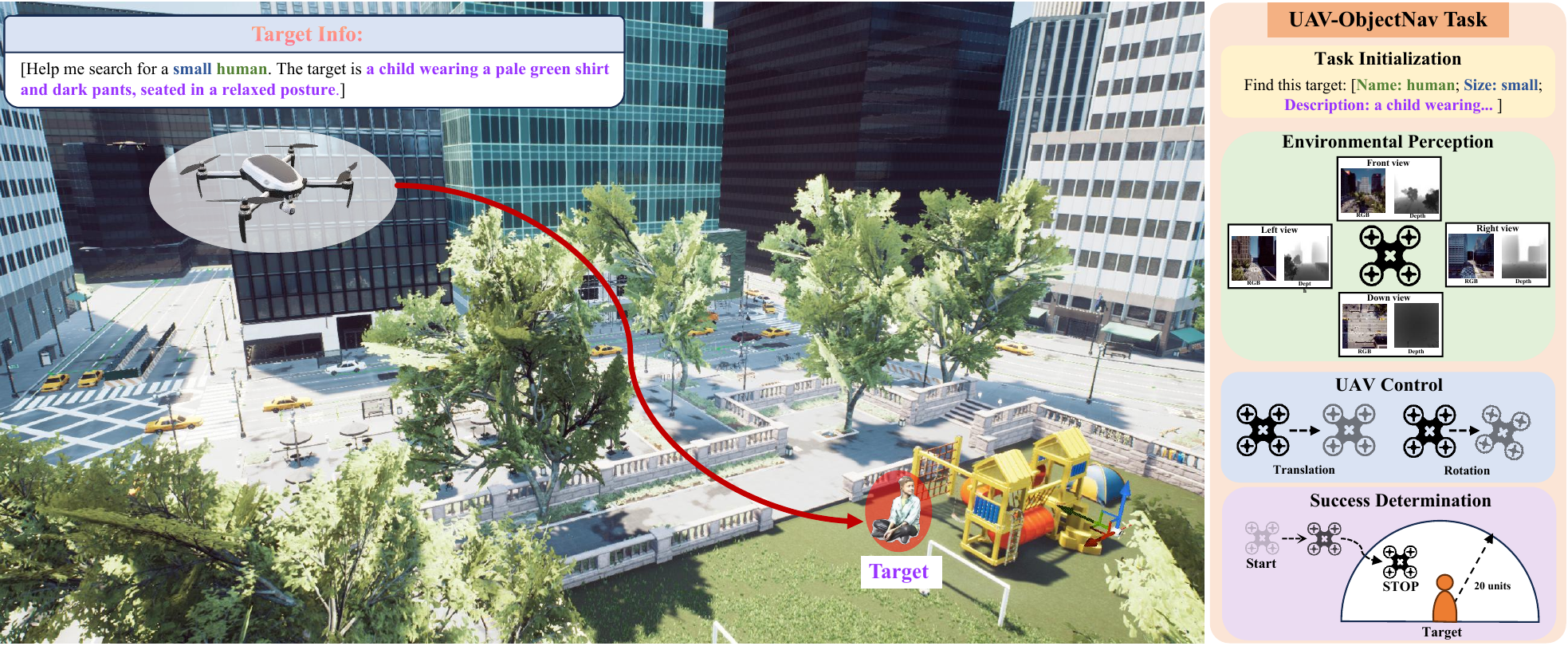}
	\caption{
		Task definition in the UAV-ON benchmark. The aerial agent receives a semantic instruction and is required to navigate in a 3D environment to locate the described target.
	}
	\label{fig:overview}
\end{figure}

\section{Task Definition}

UAV-ON defines an instance-level ObjectNav task in open-world settings, where UAVs navigate based on semantic instructions. At the beginning of each episode, the UAV is randomly initialized with a 6-DoF pose \(P_0 = [x, y, z, 0, \psi, 0]\), where \((x, y, z) \in \mathbb{R}^3\) denotes its position and \(\psi\) is the yaw angle. The agent receives a semantic instruction \(c = \{\text{name, size, description}\} \in \mathcal{C}\), which includes the target object’s category, an estimated size, and an instance-level visual description. The goal object is guaranteed to lie within a horizontal search radius \(r=50\) units centered at the starting location. With only multi-view RGB-D cameras onboard, the UAV performs navigation without GPS or global maps, relying entirely on egocentric perception. The agent is expected to understand the semantic instruction, infer object-scene relevance, and explore efficiently using only onboard sensory inputs. The action space consists of continuous, parameterized controls for translation, rotation, and stopping. Each episode terminates when the agent outputs a \texttt{Stop} command, collides with an obstacle, or reaches the maximum step limit of 150. Following~\cite{liu2023aerialvln}, a success episode is defined by the agent stopping within 20 units of the target object.

\section{UAV Simulator}
We build our benchmark using Unreal Engine~\cite{2019unrealengine} and Microsoft AirSim~\cite{shah2018airsim}, which support realistic UAV navigation and large-scale task design across diverse environments.

\subsection{Sensors for UAV}
At each timestep \(t\), the UAV obtains multi-modal observations through four synchronized RGB-D cameras oriented forward, leftward, rightward, and downward, enabling diverse perception of the environment from a limited egocentric perspective. We denote the visual input at step \( t \) as \( v_t^R = \{v_{t,\text{front}}^R, v_{t,\text{left}}^R, v_{t,\text{right}}^R, v_{t,\text{down}}^R\} \) and \( v_t^D = \{v_{t,\text{front}}^D, v_{t,\text{left}}^D, v_{t,\text{right}}^D, v_{t,\text{down}}^D\} \), where \( v_{t,*}^R \) and \( v_{t,*}^D \) represent the RGB and depth images from each corresponding direction, respectively. All views are rendered with consistent resolution and field of view, and temporally synchronized. The agent operates without access to any global localization signals, including GPS, top-down maps, or external positioning systems. It is also deprived of privileged information such as object poses, semantic maps, or underlying scene geometry. Navigation relies entirely on egocentric visual observations and an internal memory maintained over time.

\subsection{Action Space}
The task employs a parameterized action space where the agent selects from motion primitives—translation, rotation, and stopping—each associated with a continuous control parameter. Translational primitives include \texttt{Move Forward}, \texttt{Move Left}, \texttt{Move Right}, \texttt{Ascend}, and \texttt{Descend}, while rotational actions comprise \texttt{Rotate Left} and \texttt{Rotate Right}. Each selected action is defined by a corresponding distance or angular displacement, enabling the agent to dynamically adjust its movement magnitude in response to its surroundings. The \texttt{Stop} action is triggered when the agent estimates it is within 20 units of the target object and chooses to terminate the episode.

Unlike prior aerial navigation benchmarks that rely on fixed-step, discrete action spaces, this formulation provides a continuous control interface that more accurately reflects the real-world dynamics of UAVs. Actions are physically executed rather than teleported, requiring agents to ensure each selected motion yields a collision-free trajectory. Any contact with obstacles is considered a failure, significantly raising the standards for safe and reliable navigation. The setup supports fixed-step motion strategies~\cite{liu2023aerialvln} while also enabling adaptive, context-sensitive behaviors in challenging environments with clutter or semantic ambiguity. With fine-grained motion control, the agent can more precisely adapt to variations in object proximity, obstacle density, and goal ambiguity. Its pose is continuously adjusted according to the semantics of the chosen action and its control magnitude, enabling seamless integration between perception and action.

\section{UAV-ON Benchmark}
\subsection{Scene Construction}
UAV-ON consists of high-fidelity outdoor environments developed in Unreal Engine, capturing a broad spectrum of natural and artificial landscapes. All environments are segmented into regions labeled with semantic categories, including village, town, city, park, road, forest, mountain, snowy mountain, and water area. These categories reflect diverse real-world geographies, offering substantial variation in both environmental layout and navigational context. Spatial properties such as elevation variation, obstacle density, and open space ratio differ significantly across all scenes, affecting navigation difficulty and the design of effective movement strategies.

To assign contextually meaningful navigation targets, we adopt a prompt-based object mapping strategy informed by real-world co-occurrence priors. For each region type, a large language model (LLM) is queried with scene-specific prompts to generate candidate objects that are likely to appear in that setting. The generated results are manually filtered to ensure semantic relevance and spatial plausibility. Ambiguous or visually indistinct categories are removed, and selected objects are placed at navigable, context-appropriate locations. For example, benches and trash bins are located in parks; bicycles are placed along roads; and boats are often found near water areas. This co-occurrence-guided placement encourages agents to reason about object-location relationships rather than memorize fixed positions. Full prompt templates and region-to-object mappings will be released alongside the dataset.

\subsection{Dataset Analysis}
\begin{figure}[!t]
	\centering
	\includegraphics[width=0.8\linewidth]{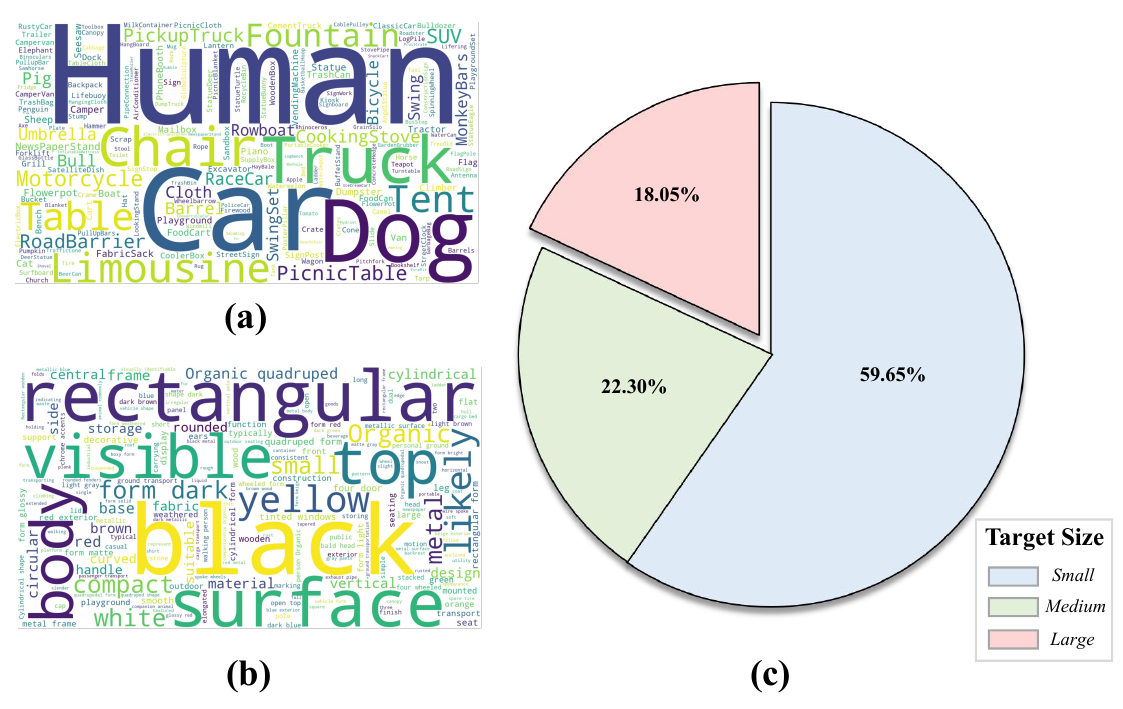}
	\caption{Statistics of target semantic information. (a) Word cloud of object category names. (b) Word cloud of descriptive attributes. (c) Distribution of target object sizes.}
	\label{fig:wordcloud}
\end{figure}
UAV-ON includes 14 high-fidelity outdoor environments with diverse geographic structures, ranging from compact urban parks to expansive mountainous terrains. Each scene contains multiple semantically labeled regions and varies in horizontal scale (e.g., from $350 \times 250$ to $1400 \times 1250$ units), resulting in rich spatial and structural diversity for aerial navigation. Across all scenes, we place a total of 1270 unique target objects over a total horizontal area of approximately 9 million square units, yielding an average object density of about 1.4 per 100 square units. Each object is paired with a semantic instruction that enables agents to reason about both appearance and contextual relevance. To illustrate the variety of target categories and linguistic expressions, we generate word clouds from the name and description fields (see Figure~\ref{fig:wordcloud}), which highlight the breadth of concepts and cues used to guide navigation. The integration of realistic, large-scale environments with semantically grounded targets and linguistically rich instructions makes UAV-ON a challenging and comprehensive benchmark for aerial ObjectNav.

\subsection{Dataset Split}
We construct the training set using 10 diverse outdoor environments, generating a total of 10,000 navigation episodes. The number of tasks per environment is proportionally allocated based on spatial size and object distribution to ensure balanced coverage across different scene types. Each episode is initialized at a random location and paired with a reachable target within a fixed search radius, forming a local navigation context. During training, the UAV can access its absolute position and the real-time Euclidean distance to the target, allowing reinforcement learning ~\cite{sutton1998reinforcement} methods to optimize navigation strategies based on egocentric observations and distance-based rewards. Furthermore, we voxelize each scene into 1-unit 3D grids and apply a three-dimensional A* algorithm ~\cite{Hart1968} to compute the shortest traversable path from the agent to the target. These paths provide safe, expert trajectories that can be used to support imitation learning approaches ~\cite{schaal1996learning}. Our training environments form a semantically diverse and spatially rich search space, offering strong support for both interaction-based and demonstration-based learning methods.

For evaluation, we collect a total of 1,000 test episodes across both the 10 training environments and 4 additional environments. Each episode is defined by a random starting position and a goal object located within a fixed local search radius. To assess generalization, the test set comprises a mix of familiar and novel scenes, as well as various object categories. In addition, several targets in the training environments are replaced with unseen categories, while the held-out environments include both reused and entirely new object categories. This setup ensures that agents are evaluated under varying levels of environmental and semantic novelty, supporting a comprehensive assessment of generalization performance.
\subsection{Metrics}
We evaluate agent performance using four metrics: Success Rate (SR), Oracle Success Rate (OSR), Distance to Success (DTS), and Success-weighted Path Length (SPL). These metrics are widely used in embodied ObjectNav tasks and are adapted for 3D aerial navigation settings. Together, they provide a comprehensive assessment of the agent’s ability to reason over complex outdoor environments, navigate efficiently, and ultimately reach target objects.

\textbf{Success Rate (SR)} measures whether the agent successfully stops within a threshold distance of the goal object at the end of each episode. It is formally defined as:
\begin{equation}
	\text{SR} = \frac{1}{N} \sum_{i=1}^{N} \mathbbm{1}{\{d_i \leq \tau\}},
	\label{eq:sr}
\end{equation}
where \( \mathbbm{1}{\{d_i \leq \tau\}} \) is an indicator function that returns 1 if the final distance \( d_i \) between the agent and the target in episode \( i \) is less than or equal to the success threshold \( \tau = 20 \). Higher SR values indicate better task completion under spatial constraints. 

\textbf{Oracle Success Rate (OSR)} measures whether the agent has ever come within the success threshold during the episode, regardless of where it stopped. This metric reflects the agent’s capacity to explore and reach near the goal, even if it fails to issue the correct Stop action:
\begin{equation}
	\text{OSR} = \frac{1}{N} \sum_{i=1}^{N} \mathbbm{1}{\left\{\min_t d_{i,t} \leq \tau \right\}},
	\label{eq:osr}
\end{equation}
where \( d_{i,t} \) is the distance to the target at timestep \( t \) in episode \( i \), and \( \tau = 20 \) is the success threshold. OSR captures whether the agent ever reaches close to the goal during its trajectory and typically serves as an upper bound to SR. 

\textbf{Distance to Success (DTS)} measures the Euclidean distance between the agent and the target at the final timestep of each episode:
\begin{equation}
	\text{DTS} = \frac{1}{N} \sum_{i=1}^{N} d_i,
	\label{eq:dts}
\end{equation}
DTS offers a continuous assessment of the agent’s final distance to the goal, complementing binary success metrics by capturing near-success behaviors. 

\textbf{Success-weighted Path Length (SPL)} jointly considers both task success and trajectory efficiency. It compares the navigation path length to the shortest possible path, while only counting successful episodes, thereby penalizing agents that reach the goal inefficiently or wander excessively before stopping:
\begin{equation}
	\text{SPL} = \frac{1}{N} \sum_{i=1}^{N} \mathbbm{1}{\{d_i \leq \tau\}} \cdot \frac{l_i}{\max(p_i, l_i)},
	\label{eq:spl}
\end{equation}
Here, \( l_i \) is the shortest geodesic distance between the start and the goal, computed by discretizing the environment into a 1-unit occupancy grid and applying the A*~\cite{Hart1968} algorithm. \( p_i \) denotes the actual trajectory length of the agent in episode \( i \).

\section{Experiments and Results}
\subsection{Baselines}
\textbf{Random:} This baseline randomly selects a movement direction at each timestep from the set of available actions, using a fixed translation and rotation step size. The agent does not utilize any observation or goal information. To prevent premature termination, the \texttt{Stop} action is disabled during the first 10 steps and becomes available thereafter. The episode terminates when the agent chooses to stop, collides with an obstacle, or reaches the maximum allowed number of steps. 

\textbf{CLIP-based Heuristic Exploration (CLIP-H):} This baseline combines CLIP-based image-text matching with a rule-based exploration strategy. At each timestep, the agent encodes four RGB views (front, left, right, and down) using a pre-trained CLIP model~\cite{radford2021learning} and compares them against the target’s textual description. If the cosine similarity between any view and the description exceeds a predefined threshold, the agent issues a \texttt{Stop} action. Otherwise, it moves in the direction with the highest similarity score using a fixed step size. To ensure meaningful and constrained navigation, we apply two post-processing rules: actions that would move the UAV beyond the predefined scene boundaries are suppressed to avoid unproductive exploration outside the task-relevant region, and a minimum flight altitude is enforced to prevent the agent from flying too close to the ground or into collision-prone zones. 
\begin{figure*}[!htbp] 
	\centering
	\includegraphics[width=\textwidth]{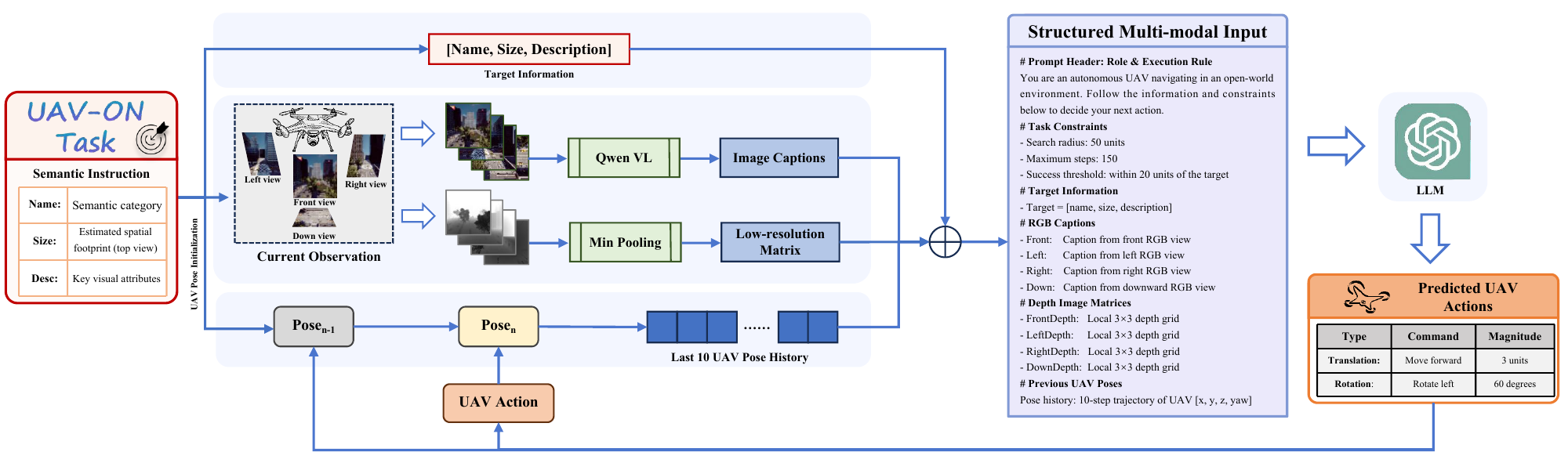}
	\caption{System overview of the Aerial ObjectNav Agent (AOA). Multi-view RGB-D observations, target information, and pose history are encoded into a structured input and processed by LLM to predict UAV actions.}
	\label{fig:baseline}
\end{figure*}
\textbf{Aerial ObjectNav Agent (AOA):} We introduce a zero-shot navigation system that taps into the spatial reasoning and exploration capabilities of LLMs within complex, open-world 3D settings. At each timestep, the agent receives RGB and depth inputs from four fixed perspectives — front, left, right, and downward — paired with a structured target prompt that describes the object’s name, estimated size, and visual characteristics. RGB frames are transformed into textual descriptions using a pre-trained Qwen-VL model~\cite{wang2024qwen2}, while depth data is distilled into compact \(3\times3 \) spatial matrices through min pooling. The system also logs the last 10 poses as 4D vectors \([x, y, z, \psi]\), offering contextual cues on recent movements.

All these multimodal inputs—including target semantics, image-language descriptions, geometric depth maps, pose history, and task constraints—are integrated into a structured prompt, which is subsequently passed to GPT-4o mini~\cite{achiam2023gpt}. Functioning as a high-level controller, GPT generates a semantic action command along with a continuous control magnitude, such as “move forward 3 units” or “rotate left 60 degrees.” To facilitate comparative evaluation, we develop two variants of the AOA framework. \textbf{AOA (Fixed-step)} employs constant translation and rotation parameters, mirroring previous UAV navigation baselines~\cite{liu2023aerialvln, lee2024citynav}, thus ensuring compatibility with discrete control benchmarks. By contrast, \textbf{AOA (Variable-step)} exploits the reasoning ability of GPT to flexibly determine both the action type and its corresponding magnitude, aligning more naturally with the benchmark’s continuous action space and allowing for greater behavioral adaptability. Both AOA variants operate entirely in a zero-shot regime without any model training or fine-tuning, offering a standardized framework for evaluating the spatial generalization capabilities of LLMs.To ensure navigational validity, we implement a post-processing mechanism that replaces out-of-bound movement actions with in-place rotations, allowing the UAV to reorient itself without exiting the scene. This adjustment was motivated by early observations that GPT occasionally exhibits limited spatial awareness. We do not enforce a minimum flight altitude; the system relies on onboard depth cameras to avoid low-altitude collisions.

\begin{table*}[!t]
	\small
	\centering
	\caption{Performance across different object sizes and overall total, evaluated using four metrics: Distance to Success (DTS↓, in simulation units), Success Rate (SR↑), Oracle Success Rate (OSR↑), and Success weighted by Path Length (SPL↑).}
	\label{tab:size_results}
	\resizebox{\textwidth}{!}{
		\begin{tabular}{lcccccccccccccccc}
			\toprule
			\multirow{2}{*}{\textbf{Method}} 
			& \multicolumn{4}{c}{\textbf{Small}} 
			& \multicolumn{4}{c}{\textbf{Medium}} 
			& \multicolumn{4}{c}{\textbf{Large}} 
			& \multicolumn{4}{c}{\textbf{Total}} \\
			& DTS↓ & SR↑ & OSR↑ & SPL↑ 
			& DTS↓ & SR↑ & OSR↑ & SPL↑ 
			& DTS↓ & SR↑ & OSR↑ & SPL↑ 
			& DTS↓ & SR↑ & OSR↑ & SPL↑ \\
			\midrule
			Random 
			& \textbf{43.08} & 4.14\% & 7.80\% & \textbf{2.80}\%
			& \textbf{42.62} & 3.33\% & 8.10\% & 3.05\%
			& 40.59 & 2.48\% & 8.07\% & 1.62\%
			& \textbf{42.57} & 3.70\% & 8.00\% & 2.66\% \\
			CLIP-H 
			& 48.64 & 2.86\% & 8.43\% & 1.51\%
			& 43.91 & \textbf{10.95}\% & 16.67\% & \textbf{7.17}\%
			& 40.38 & 13.04\% & 19.25\% & 10.53\%
			& 46.31 & 6.20\% & 11.90\% & \textbf{4.15}\% \\
			AOA-V 
			& 50.56 & 2.86\% & \textbf{25.44}\% & 0.54\%
			& 48.96 & 5.71\% & \textbf{27.62}\% & 1.73\%
			& 46.58 & 7.45\% & \textbf{27.95}\% & 1.038\%
			& 49.58 & 4.20\% & \textbf{26.30}\% & 0.87\% \\
			AOA-F 
			& 50.67 & \textbf{4.45}\% & 16.38\% &1.61\%
			& 48.14 & 10.48\% & 17.62\% & 6.36\%
			& \textbf{39.68} & 14.29\% & 21.74\% & \textbf{10.66}\%
			& 48.37 & \textbf{7.30}\% & 17.50\% & 4.06\% \\
			\bottomrule
		\end{tabular}
	}
\end{table*}

\subsection{Results}

\begin{figure}[htbp]
	\centering
	\begin{minipage}[t]{0.38\textwidth}
		As shown in Table~\ref{tab:size_results}, we evaluate four baseline methods across different object sizes using standard navigation metrics. AOA-V achieves the highest overall OSR, indicating its firm semantic grounding and ability to explore large areas of the scene effectively. This performance demonstrates its ability to gain proximity to targets, even in unfamiliar settings. However,
		
		\vspace{-3em}  
	\end{minipage}
	\hfill
	\begin{minipage}[t]{0.6\textwidth}
		\scriptsize
		\centering
		\captionof{table}{Analysis of termination behavior and safe navigation performance under different control strategies.}
		\vspace{0.5em}
		\label{tab:termination_distance}
		\begin{tabular}{lcccccc}
			\toprule
			\multirow{2}{*}{\textbf{Method}} 
			& \multicolumn{3}{c}{\textbf{Termination Type (\%)}} 
			& \multicolumn{2}{c}{\textbf{Navigation Statistics}} \\
			& Stop & Max Step & Collision & Avg. Steps & Safe Dist. (m) \\
			\midrule
			Random  & \textbf{62.1} & 0.0 & 37.9 & 12.09 & 31.68 \\
			CLIP-H  & 36.8 & 11.8 & 51.4 & 30.25 & 125.21 \\
			AOA-V   & 19.9 & \textbf{35.1} & 45.0 & \textbf{85.65} & \textbf{232.08} \\
			AOA-F   & 30.6 & 3.9 & \textbf{65.5} & 36.88 & 144.25 \\
			\bottomrule
		\end{tabular}
	\end{minipage}
\end{figure}

 its lower SR and SPL underscore a key limitation: the LLM is tasked with simultaneously handling semantic understanding, motion planning, and termination control. This multitasking can dilute attention, making it difficult to execute precise, goal-aligned stop decisions. AOA-F, by contrast, delivers more consistent performance in both SR and SPL, especially when navigating toward large and visually salient objects. Its fixed-step motion simplifies control, enabling more reliable trajectory execution and successful stops, albeit with slightly reduced exploratory reach. CLIP-H stands out in terms of SPL, suggesting efficient path following driven by strong visual similarity. However, its lower OSR and SR indicate a limited understanding of semantic goals and a reliance on static stop heuristics rather than contextual reasoning. The Random baseline, as expected, performs the worst across all metrics, often failing to establish meaningful progress toward any target. These results reveal clear trade-offs across the methods: AOA-V prioritizes semantic exploration, AOA-F emphasizes controllability and stability, and CLIP-H favors visual matching over semantic grounding.

Table~\ref{tab:termination_distance} further reveals how these methods behave in practice. AOA-V covers large areas during navigation, reflecting its strong exploration capability, but often fails to terminate effectively, underscoring control instability when precise stopping is required. AOA-F presents a more balanced behavior, combining purposeful movement with more effective use of the \texttt{Stop} action, though occasional control failures remain, particularly near obstacles. CLIP-H displays greater stop reliability and fewer erratic terminations, thanks to its reliance on strong visual anchors for action selection, though it explores less aggressively. The Random baseline halts without intent, further confirming its lack of goal understanding. Notably, all methods exhibit collision rates exceeding 30\%, which would be unacceptable in real-world UAV deployments where safety is critical. Such high collision rates pose serious risks in physical environments, where hardware damage and human safety must be considered. This highlights a significant gap between current navigation policies and the requirements of real-world aerial systems. We advocate the use of our UAV-ON benchmark as a platform to develop safer and more robust control strategies for aerial embodied agents operating in open-world environments.

\section{Conclusion}
We present UAV-ON, a large-scale, semantically grounded benchmark designed for object-goal navigation by UAVs in open-world 3D environments. The benchmark comprises 14 high-fidelity outdoor scenes that span diverse geographic and structural layouts and defines thousands of navigation tasks based on structured prompts that encode object categories, physical dimensions, and visual features. To assess UAV-ON, we implement a vision-language-based agent and perform systematic zero-shot evaluations. The results underscore the compounded challenges of semantic reasoning, obstacle-aware exploration, and target localization in aerial navigation contexts. We envision UAV-ON as a foundation for advancing research in multimodal perception, prompt-conditioned control, and scalable autonomy for UAVs operating in complex real-world environments.

\bibliographystyle{plainnat}
\bibliography{sample-base-arxiv}

@String{Computing = "Computing" }

@String{Computer = "{IEEE} Computer" }

@String{Springer = "Springer-Verlag" }

@incollection{golabi2022,
  title={Intelligent and Fuzzy UAV Transportation Applications in Aviation 4.0},
  author={Golabi, Mahmoud and Nejad, Mazyar Ghadiri},
  booktitle={Intelligent and Fuzzy Techniques in Aviation 4.0},
  series={Studies in Systems, Decision and Control},
  volume={372},
  pages={431--458},
  year={2022},
  publisher={Springer, Cham},
  doi={10.1007/978-3-030-75067-1_19}
}

@article{barmpounakis2016,
  title={Unmanned Aerial Aircraft Systems for transportation engineering: Current practice and future challenges},
  author={Barmpounakis, Emmanouil N. and Vlahogianni, Eleni I. and Golias, John C.},
  journal={International Journal of Transportation Science and Technology},
  volume={5},
  number={3},
  pages={111--122},
  year={2016},
  publisher={Elsevier}
}

@article{zhao2024,
  title={Joint optimization of loading, mission abort and rescue site selection policies for UAV},
  author={Zhao, Xian and Wang, Xinlei and Dai, Ying and Qiu, Qingan},
  journal={Reliability Engineering \& System Safety},
  volume={244},
  pages={109955},
  year={2024},
  publisher={Elsevier},
  doi={10.1016/j.ress.2024.109955}
}

@inproceedings{scherer2015,
  title={An autonomous multi-UAV system for search and rescue},
  author={Scherer, J. and Yahyanejad, S. and Hayat, S. and others},
  booktitle={Proceedings of the first workshop on micro aerial vehicle networks, systems, and applications for civilian use},
  pages={33--38},
  year={2015}
}

@article{asadzadeh2022,
  title={UAV-based remote sensing for the petroleum industry and environmental monitoring: State-of-the-art and perspectives},
  author={Asadzadeh, Saeid and de Oliveira, Wilson José and de Souza Filho, Carlos Roberto},
  journal={Journal of Petroleum Science and Engineering},
  volume={208},
  pages={109633},
  year={2022},
  publisher={Elsevier}
}

@article{liu2022,
  title={UAV trajectory optimization for time-constrained data collection in UAV-enabled environmental monitoring systems},
  author={Liu, Kai and Zheng, Jun},
  journal={IEEE Internet of Things Journal},
  volume={9},
  number={23},
  pages={24300--24314},
  year={2022},
  publisher={IEEE}
}

@article{mohamed2020,
  title={Unmanned aerial vehicles applications in future smart cities},
  author={Mohamed, Nader and Al-Jaroodi, Jameela and Jawhar, Imad and Mohamed, Ferial and Mahmoud, Sameh},
  journal={Technological Forecasting and Social Change},
  volume={153},
  pages={119293},
  year={2020},
  publisher={Elsevier},
  doi={10.1016/j.techfore.2020.119293}
}

@article{pongsakornsathien2025advances,
  title={Advances in low-altitude airspace management for uncrewed aircraft and advanced air mobility},
  author={Pongsakornsathien, Nichakorn and Safwat, Nour El-Din and Xie, Yibing and Gardi, Alessandro and Sabatini, Roberto},
  journal={Progress in Aerospace Sciences},
  pages={101085},
  year={2025},
  publisher={Elsevier}
}

@article{wang2020,
  title={UAV environmental perception and autonomous obstacle avoidance: A deep learning and depth camera combined solution},
  author={Wang, Dashuai and Li, Wei and Liu, Xiaoguang and Li, Nan and Zhang, Chunlong},
  journal={Computers and Electronics in Agriculture},
  volume={175},
  pages={105523},
  year={2020},
  publisher={Elsevier},
  doi={10.1016/j.compag.2020.105523}
}

@article{bijjahalli2020,
  title={Advances in intelligent and autonomous navigation systems for small UAS},
  author={Bijjahalli, Suraj and Sabatini, Roberto and Gardi, Alessandro},
  journal={Progress in Aerospace Sciences},
  volume={115},
  pages={100617},
  year={2020},
  publisher={Elsevier},
  doi={10.1016/j.paerosci.2020.100617}
}

@inproceedings{chu2024,
  title={Towards Natural Language-Guided Drones: GeoText-1652 Benchmark with Spatial Relation Matching},
  author={Chu, Meng and Zheng, Zhedong and Ji, Wei and Wang, Tingyu and Chua, Tat-Seng},
  booktitle={European Conference on Computer Vision (ECCV)},
  pages={213--231},
  year={2024},
  publisher={Springer Nature Switzerland}
}

@article{wang2024,
  title={Towards Realistic UAV Vision-Language Navigation: Platform, Benchmark, and Methodology},
  author = {Wang, Xiangyu and Yang, Donglin and Wang, Ziqin and Kwan, Hohin and others},
  journal={arXiv preprint arXiv:2410.07087},
  year={2024}
}

@article{gao2025,
  title={OpenFly: A Versatile Toolchain and Large-scale Benchmark for Aerial Vision-Language Navigation},
  author={Gao, Yunpeng and Li, Chenhui and You, Zhongrui and Liu, Junli and others},
  journal={arXiv preprint arXiv:2502.18041},
  year={2025}
}

@article{zhao2025,
  title={Aerial Vision-and-Language Navigation with Grid-based View Selection and Map Construction},
  author={Zhao, Ganlong and Li, Guanbin and Pan, Jia and Yu, Yizhou},
  journal={arXiv preprint arXiv:2503.11091},
  year={2025}
}

@article{hong2025gsavln,
  title={General Scene Adaptation for Vision-and-Language Navigation},
  author={Hong, Haodong and Qiao, Yanyuan and Wang, Sen and Liu, Jiajun and Wu, Qi},
  journal={arXiv preprint arXiv:2501.17403},
  year={2025}
}

@article{anderson2018evaluation,
  title={On evaluation of embodied navigation agents},
  author={Anderson, Peter and Chang, Angel and Chaplot, Devendra Singh and Dosovitskiy, Alexey and Gupta, Saurabh and Koltun, Vladlen and Kosecka, Jana and Malik, Jitendra and Mottaghi, Roozbeh and Savva, Manolis and others},
  journal={arXiv preprint arXiv:1807.06757},
  year={2018}
}

@article{kolve2017ai2thor,
  title={Ai2-thor: An interactive 3d environment for visual ai},
  author={Kolve, Eric and Mottaghi, Roozbeh and Han, Winson and others},
  journal={arXiv preprint arXiv:1712.05474},
  year={2017}
}

@inproceedings{xia2018gibson,
  title={Gibson env: Real-world perception for embodied agents},
  author={Xia, Fei and Zamir, Amir R. and He, Zhiyang and Sax, Alexander and others},
  booktitle={Proceedings of the IEEE/CVF Conference on Computer Vision and Pattern Recognition},
  pages={9068--9079},
  year={2018}
}

@inproceedings{deitke2020robothor,
  title={Robothor: An open simulation-to-real embodied ai platform},
  author={Deitke, Matt and Han, Winson and Herrasti, Alvaro and Kembhavi, Aniruddha and others},
  booktitle={Proceedings of the IEEE/CVF Conference on Computer Vision and Pattern Recognition},
  pages={3164--3174},
  year={2020}
}

@article{ramakrishnan2021hm3d,
  title={Habitat-Matterport 3D dataset (HM3D): 1000 large-scale 3D environments for embodied AI},
  author={Ramakrishnan, Santhosh K. and Gokaslan, Aaron and Wijmans, Erik and Maksymets, Oleksandr and others},
  journal={arXiv preprint arXiv:2109.08238},
  year={2021}
}

@article{ye2021hiem,
  title={Efficient robotic object search via hiem: Hierarchical policy learning with intrinsic-extrinsic modeling},
  author={Ye, Xin and Yang, Yezhou},
  journal={IEEE Robotics and Automation Letters},
  volume={6},
  number={3},
  pages={4425--4432},
  year={2021},
  publisher={IEEE}
}

@inproceedings{ye2021auxiliary,
  title={Auxiliary tasks and exploration enable objectgoal navigation},
  author={Ye, Joel and Batra, Dhruv and Das, Abhishek and Wijmans, Erik},
  booktitle={Proceedings of the IEEE/CVF International Conference on Computer Vision},
  pages={16117--16126},
  year={2021}
}

@inproceedings{ramrakhya2023pirlnav,
  title={Pirlnav: Pretraining with imitation and rl finetuning for objectnav},
  author={Ramrakhya, Ram and Batra, Dhruv and Wijmans, Erik and Das, Abhishek},
  booktitle={Proceedings of the IEEE/CVF Conference on Computer Vision and Pattern Recognition},
  pages={17896--17906},
  year={2023}
}

@article{chaplot2020objectgoal,
  title={Object goal navigation using goal-oriented semantic exploration},
  author={Chaplot, Devendra Singh and Gandhi, Dhiraj Prakashchand and Gupta, Abhinav and Salakhutdinov, Ruslan},
  journal={Advances in Neural Information Processing Systems},
  volume={33},
  pages={4247--4258},
  year={2020}
}

@article{gao2024stmr,
  title={Aerial Vision-and-Language Navigation via Semantic-Topo-Metric Representation Guided LLM Reasoning},
  author={Gao, Yunpeng and Wang, Zhigang and Han, Pengfei and Jing, Linglin and others},
  journal={arXiv preprint arXiv:2410.08500},
  year={2024}
}

@inproceedings{luo2022stubborn,
  title={Stubborn: A strong baseline for indoor object navigation},
  author={Luo, Haokuan and Yue, Albert and Hong, Zhang-Wei and Agrawal, Pulkit},
  booktitle={2022 IEEE/RSJ International Conference on Intelligent Robots and Systems (IROS)},
  pages={3287--3293},
  year={2022},
  organization={IEEE}
}

@inproceedings{campari2022abstractmodels,
  title={Online learning of reusable abstract models for object goal navigation},
  author={Campari, Tommaso and Lamanna, Leonardo and Traverso, Paolo and Serafini, Luciano and Ballan, Lamberto},
  booktitle={Proceedings of the IEEE/CVF Conference on Computer Vision and Pattern Recognition},
  pages={14870--14879},
  year={2022}
}

@inproceedings{zhao2023zeroshot,
  title={Zero-shot object goal visual navigation},
  author={Zhao, Qianfan and Zhang, Lu and He, Bin and Qiao, Hong and Liu, Zhiyong},
  journal={arXiv preprint arXiv:2206.07423},
  year={2022}
}

@article{zhao2023semantic,
  title={Semantic policy network for zero-shot object goal visual navigation},
  author={Zhao, Qianfan and Zhang, Lu and He, Bin and Liu, Zhiyong},
  journal={IEEE Robotics and Automation Letters},
  volume={8},
  number={11},
  pages={7655--7662},
  year={2023},
  publisher={IEEE}
}

@article{chen2023zeroshot,
  title={Zero-shot object searching using large-scale object relationship prior},
  author={Chen, Hongyi and Xu, Ruinian and Cheng, Shuo and Vela, Patricio A. and Xu, Danfei},
  journal={arXiv preprint arXiv:2303.06228},
  year={2023}
}

@article{lu2018survey,
  title={A survey on vision-based UAV navigation},
  author={Lu, Yuncheng and Xue, Zhucun and Xia, Gui-Song and Zhang, Liangpei},
  journal={Geo-spatial Information Science},
  volume={21},
  number={1},
  pages={21--32},
  year={2018},
  publisher={Taylor \& Francis}
}

@article{aitsaadi2022uav,
  title={UAV path planning using optimization approaches: A survey},
  author={Ait Saadi, Amylia and Soukane, Assia and Meraihi, Yassine and Benmessaoud Gabis, Asma and Mirjalili, Seyedali and Ramdane-Cherif, Amar},
  journal={Archives of Computational Methods in Engineering},
  volume={29},
  number={6},
  pages={4233--4284},
  year={2022},
  publisher={Springer}
}

@article{jones2023path,
  title={Path-planning for unmanned aerial vehicles with environment complexity considerations: A survey},
  author={Jones, Michael and Djahel, Soufiene and Welsh, Kristopher},
  journal={ACM Computing Surveys (CSUR)},
  volume={55},
  number={11},
  pages={1--39},
  year={2023},
  publisher={ACM}
}

@inproceedings{liu2023aerialvln,
  title={AerialVLN: Vision-and-language navigation for UAVs},
  author={Liu, Shubo and Zhang, Hongsheng and Qi, Yuankai and Wang, Peng and Zhang, Yanning and Wu, Qi},
  booktitle={Proceedings of the IEEE/CVF International Conference on Computer Vision},
  pages={15384--15394},
  year={2023}
}

@article{lee2024citynav,
  title={CityNav: Language-Goal Aerial Navigation Dataset with Geographic Information},
  author={Lee, Jungdae and Miyanishi, Taiki and Kurita, Shuhei and Sakamoto, Koya and others},
  journal={arXiv preprint arXiv:2406.14240},
  year={2024}
}

@article{liu2024navagent,
  title={NavAgent: Multi-scale Urban Street View Fusion For UAV Embodied Vision-and-Language Navigation},
  author={Liu, Youzhi and Yao, Fanglong and Yue, Yuanchang and Xu, Guangluan and others},
  journal={arXiv preprint arXiv:2411.08579},
  year={2024}
}

@article{xu2025geonav,
  title={GeoNav: Empowering MLLMs with Explicit Geospatial Reasoning Abilities for Language-Goal Aerial Navigation},
  author={Xu, Haotian and Hu, Yue and Gao, Chen and Zhu, Zhengqiu and others},
  journal={arXiv preprint arXiv:2504.09587},
  year={2025}
}

@software{2019unrealengine,
  author = {{Epic Games}},
  title = {Unreal Engine},
  url = {https://www.unrealengine.com},
  version = {4.22.1},
  date = {2019-04-25},
}

@inproceedings{shah2018airsim,
  title={Airsim: High-fidelity visual and physical simulation for autonomous vehicles},
  author={Shah, Shital and Dey, Debadeepta and Lovett, Chris and Kapoor, Ashish},
  booktitle={Field and Service Robotics: Results of the 11th International Conference},
  pages={621--635},
  year={2018},
  organization={Springer}
}

@book{sutton1998reinforcement,
  title={Reinforcement learning: An introduction},
  author={Sutton, Richard S. and Barto, Andrew G.},
  volume={1},
  number={1},
  year={1998},
  publisher={MIT press Cambridge}
}

@article{schaal1996learning,
  title={Learning from demonstration},
  author={Schaal, Stefan},
  journal={Advances in neural information processing systems},
  volume={9},
  year={1996}
}

@article{Hart1968,
  doi = {10.1109/tssc.1968.300136},
  url = {https://doi.org/10.1109/tssc.1968.300136},
  year = {1968},
  publisher = {Institute of Electrical and Electronics Engineers ({IEEE})},
  volume = {4},
  number = {2},
  pages = {100--107},
  author={Hart, Peter E. and Nilsson, Nils J. and Raphael, Bertram},
  title = {A Formal Basis for the Heuristic Determination of Minimum Cost Paths},
  journal = {{IEEE} Transactions on Systems Science and Cybernetics}
}

@inproceedings{radford2021learning,
  title={Learning transferable visual models from natural language supervision},
  author={Radford, Alec and Kim, Jong Wook and Hallacy, Chris and Ramesh, Aditya and others},
  pages={8748--8763},
  year={2021},
  organization={PmLR}
}

@article{wang2024qwen2,
  title={Qwen2-vl: Enhancing vision-language model's perception of the world at any resolution},
  author={Wang, Peng and Bai, Shuai and Tan, Sinan and Wang, Shijie and others},
  journal={arXiv preprint arXiv:2409.12191},
  year={2024}
}

@article{achiam2023gpt,
  title={Gpt-4 technical report},
  author={OpenAI and Achiam, Josh and Adler, Steven and Agarwal, Sandhini and others},
  journal={arXiv preprint arXiv:2303.08774},
  year={2023}
}

@article{yue2024safe,
  title={Safe-vln: Collision avoidance for vision-and-language navigation of autonomous robots operating in continuous environments},
  author={Yue, Lu and Zhou, Dongliang and Xie, Liang and Zhang, Feitian and others},
  journal={IEEE Robotics and Automation Letters},
  year={2024},
  publisher={IEEE}
}

\newpage

\appendix
\section*{Appendix}
\addcontentsline{toc}{section}{Appendix}

\section{Prompt-Driven Scene-Object Co-occurrence Mapping}
\begin{figure}[!th]
	\centering
	\includegraphics[width=1\linewidth]{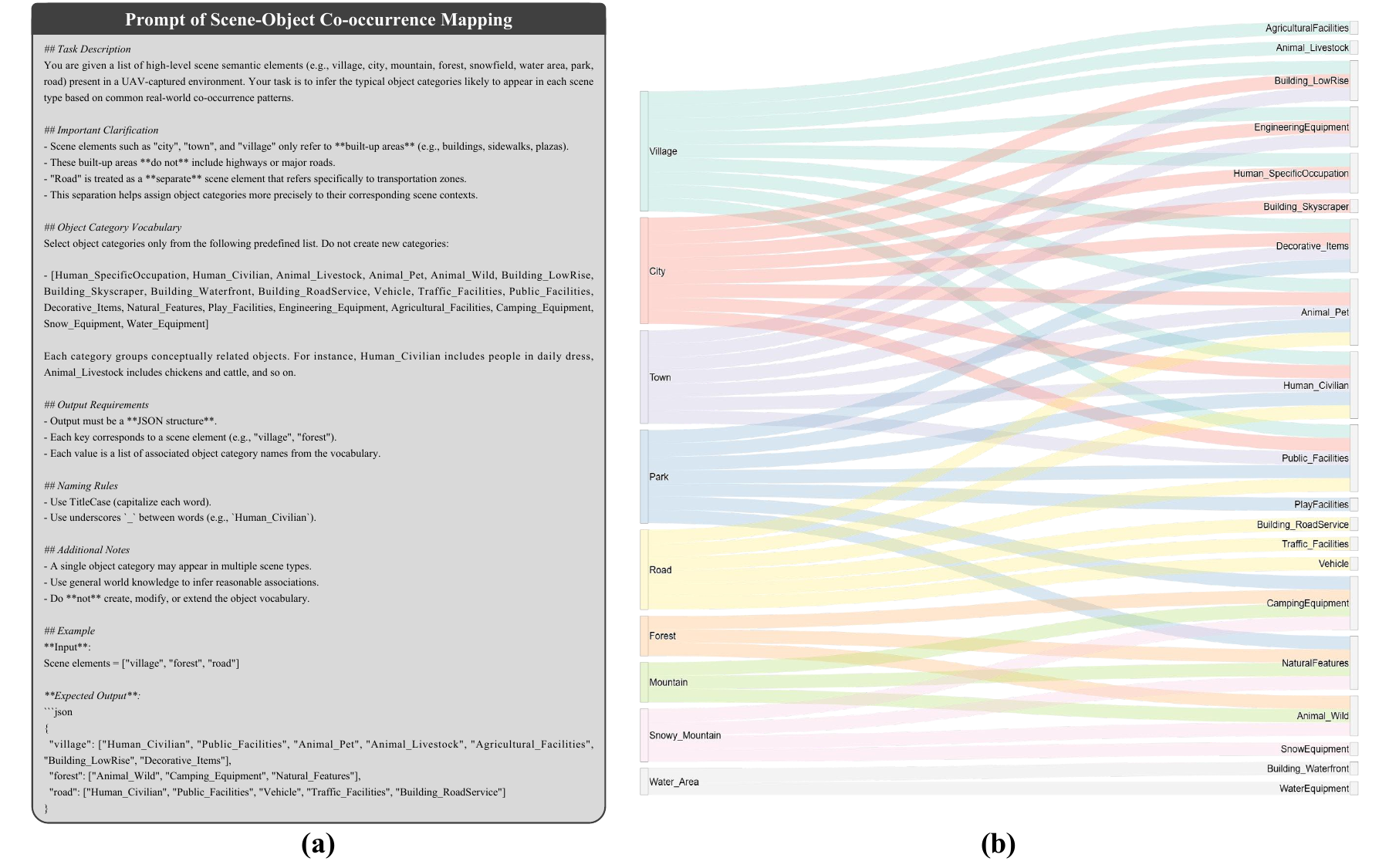}
	\caption{ (a) Task prompt used for scene-object co-occurrence inference. (b) Co-occurrence mapping visualized as a Sankey diagram linking scenes to likely object categories.}
	\label{fig:scene_object}
\end{figure}

To model semantic co-occurrence between scenes and object categories in UAV-captured environments, we propose a prompt-based knowledge extraction method using the GPT-4o-mini-high model. The goal is to establish a mapping from high-level scene semantics (e.g., \textit{village}, \textit{forest}, \textit{road}) to object categories that are commonly found in those settings. As shown in Figure~\ref{fig:scene_object}(a), the prompt is carefully designed with clear task instructions, a constrained object vocabulary, strict naming rules, and explicit formatting requirements. It also includes scene-type clarifications (e.g., distinguishing built-up areas from natural terrains) to reduce ambiguity. This structured design ensures that outputs remain consistent, interpretable, and aligned with realistic environmental priors.

The resulting scene-object mapping, visualized as a Sankey diagram in Figure~\ref{fig:scene_object}(b), captures meaningful co-occurrence patterns that mirror everyday spatial contexts. For example, \textit{village} scenes frequently co-occur with categories such as \textit{Human\_Civilian}, \textit{Animal\_Livestock}, and \textit{Agricultural\_Facilities}, while \textit{road} scenes are strongly linked to \textit{Vehicle} and \textit{Traffic\_Facilities}. These mappings serve two key purposes in UAV-ON: (1) guiding object placement to maintain contextual realism during environment generation, and (2) supporting semantic grounding in goal prompts to help agents interpret high-level navigation instructions. All GPT-generated outputs were manually reviewed to correct inconsistencies and ensure full compliance with prompt constraints, resulting in a high-quality semantic knowledge base that connects language-driven priors with physical scene design.

\begin{figure}[!t]
	\centering
	\includegraphics[width=0.8\linewidth]{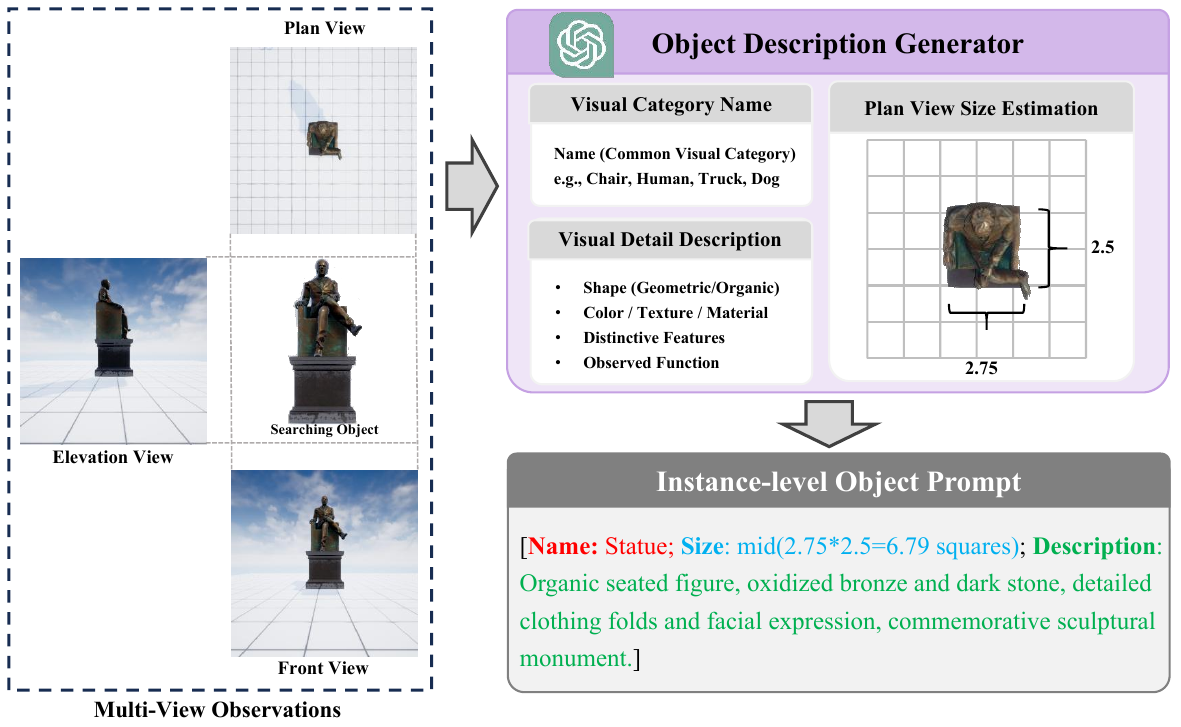}
	\caption{Illustration of the object prompt generation process. Multi-view images are used to generate an instance-level object prompt via GPT-based object description and size estimation.}
	\label{fig:gpt_description}
\end{figure}
\section{Instance-level Object Prompt Construction}
To support precise and semantically grounded object-goal navigation, we construct structured instance-level prompts that provide detailed natural language descriptions of target objects. These prompts are used to communicate high-level task goals to the navigation agent in a format compatible with large language models.
As illustrated in Figure~\ref{fig:gpt_description}, the prompt construction process begins with multi-view observations of each object instance, captured within the simulation environment. Specifically, we collect front, plan, and elevation views, which together provide comprehensive visual and spatial context. This multi-view input helps capture key features such as silhouette, dimensions, and placement relative to the ground plane—elements critical for downstream recognition and localization.
We then leverage GPT-4 to generate descriptive prompts based on these visual observations. Each prompt consists of three components: (1) a visual category label (e.g., \textit{Statue}, \textit{Chair}), (2) an estimated size in plan view (e.g., \textit{mid(2.75$\times$2.5=6.79 unit squares)}), and (3) a fine-grained natural language description that captures the object's shape, material, and semantic attributes (e.g., \textit{Organic seated figure, oxidized bronze and dark stone, commemorative sculptural monument}). These detailed instance-level prompts are then provided to the navigation model as the semantic goal input.
This approach ensures that the agent receives a rich, human-interpretable description that supports goal understanding, spatial reasoning, and object disambiguation. To facilitate reproducibility and scalability, we release a lightweight, open-source tool for automated prompt generation. Given input images, the tool formats them appropriately and interfaces with GPT to produce structured object descriptions. The toolkit and usage instructions are publicly available at: \url{https://chatgpt.com/g/g-6819b2befeac81918550152db018a5dd-object-description-generator}.

\begin{figure}[!t]
	\centering
	\includegraphics[width=1\linewidth]{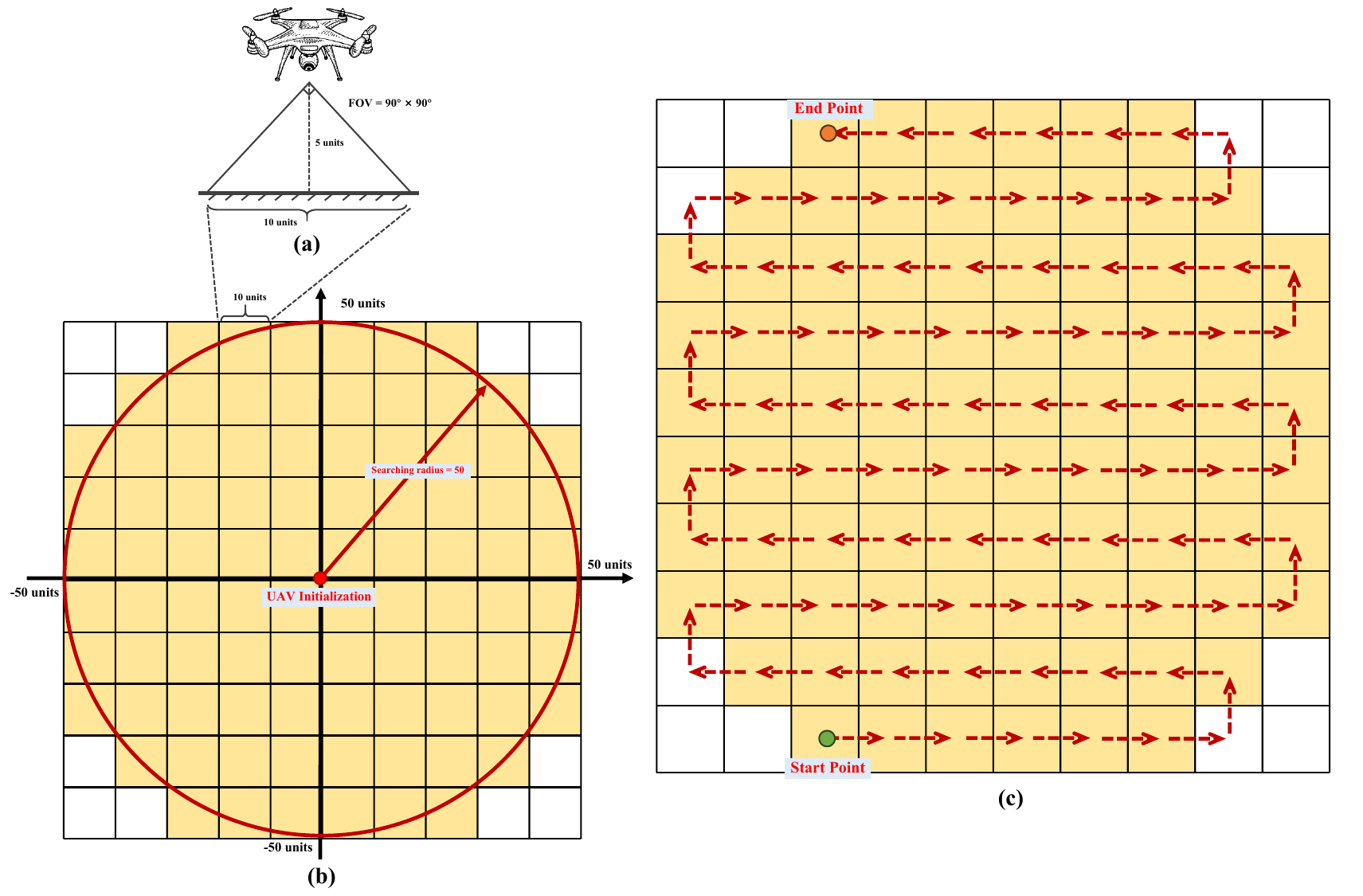}
	\caption{Geometric basis for determining the maximum exploration step limit. 
		(a) Visualization of the UAV’s downward-facing FOV at 5 meters altitude; 
		(b) Grid-based search region constrained within a 50-unit radius circle centered at the initialization point; 
		(c) Zig-zag traversal policy covering all 88 valid grid cells.}
	\label{fig:move_count}
\end{figure}
\section{Justification for the Maximum Exploration Step Limit}
To define a principled upper bound on the number of exploration steps in the UAV-ON task, we analyze the perceptual geometry of the UAV and the spatial extent of the search region. Each UAV is equipped with four RGB-D cameras, each with a field of view (FOV) of $90^\circ \times 90^\circ$. For typical small to medium-sized objects, we assume that the downward-facing camera provides sufficient coverage when the UAV hovers at an altitude of 5 meters, enabling reliable ground-level visibility without excessive occlusion (see Figure~\ref{fig:move_count}(a)).
Given this sensor configuration, we construct a discrete 2D grid map over the search region, where each cell approximates the area observable in a single downward view from a distance of 5 meters. The search region is defined as a circular area with a radius of 50 units in the $X-Y$ plane, centered at the UAV’s starting location. As visualized in Figure~\ref{fig:move_count}(b), all grid cells whose centers or partial areas fall within this circle are marked as part of the valid search space. This results in a total of 88 grid cells to be visited for complete coverage of the region.
To estimate a lower bound on the number of steps required for full exploration, we simulate a deterministic zig-zag traversal strategy using only discrete translational actions: \texttt{move forward}, \texttt{move left}, and \texttt{move right}. This policy does not utilize global planning or semantic priors, but rather reflects a basic spatial sweep approach. As shown in Figure~\ref{fig:move_count}(c), visiting all 88 valid grid cells requires 92 steps under idealized conditions. To account for practical factors such as the initial movement from the center, minor altitude adjustments, and occasional maneuvers for obstacle avoidance, we conservatively increase this estimate by a factor of 1.5. Based on these considerations, we set the maximum exploration step limit to 150. This allows sufficient action budget for agents to conduct exhaustive local exploration, while also accommodating more reactive or adaptive behaviors. Importantly, this upper bound ensures a fair evaluation setting across methods by providing enough flexibility to complete the task without artificially limiting navigation capacity.

\begin{figure}[!t]
	\centering
	\includegraphics[width=1\linewidth]{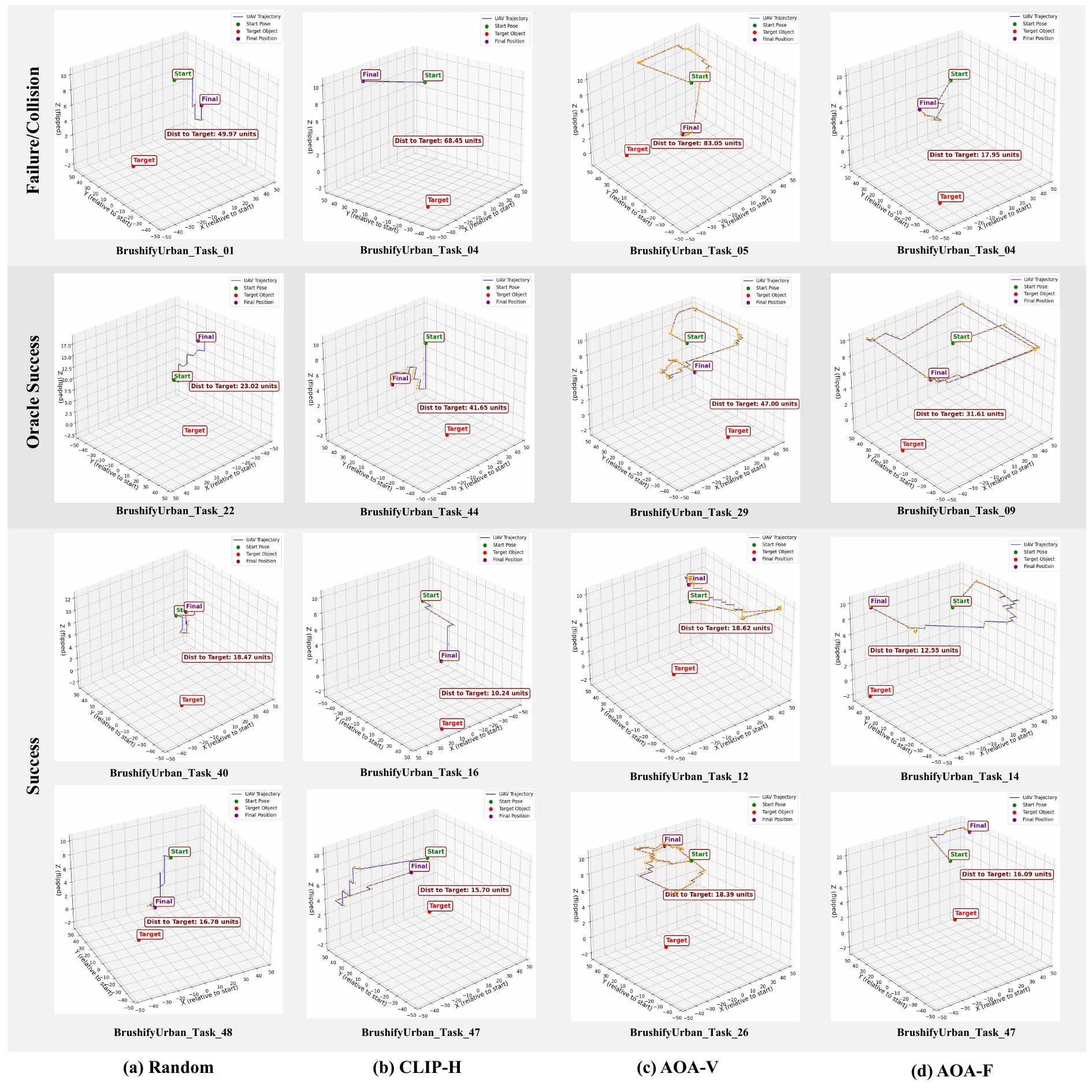}
	\caption{Representative 3D trajectories of four baseline methods across three outcome types: \textit{Failure/Collision} (top row), \textit{Oracle Success} (middle row), and \textit{Success} (bottom row). Methods include: (a) Random, (b) CLIP-H, (c) AOA-V, and (d) AOA-F. All examples are drawn from the same scene (\textit{BrushifyUrban}) for comparability.}
	\label{fig:baseline_trajectories}
\end{figure}
\section{Trajectories for Baseline Comparison}
To qualitatively assess the behavioral tendencies and control characteristics of different baseline policies, we visualize representative 3D trajectories sampled from the \textit{BrushifyUrban} environment, categorized by task outcomes: \textit{Failure/Collision}, \textit{Oracle Success}, and \textit{Success}. As shown in Figure~\ref{fig:baseline_trajectories}, each row corresponds to a distinct outcome category, while each column depicts a different method: (a) Random, (b) CLIP-H, (c) AOA-V, and (d) AOA-F.
Each subplot illustrates the UAV’s flight trajectory as a colored line, along with markers denoting the start point, termination point, and ground-truth object location. Final distances to the target are annotated to quantify spatial proximity upon termination. By sampling all trajectories from the same environment, we ensure a controlled setting for fair cross-method comparison.
The visualization reveals clear differences in policy behavior. AOA-F demonstrates relatively direct and stable navigation in successful cases, suggesting effective alignment between semantic goal prompts and trajectory planning. In contrast, AOA-V displays broader exploratory behavior, often reaching the target’s vicinity but failing to issue a timely \texttt{Stop} command, indicative of less stable control execution under variable action magnitudes. CLIP-H tends to follow concise, semantically plausible paths but occasionally terminates prematurely due to its rigid stopping logic. The Random policy exhibits erratic motion and fails to demonstrate any meaningful goal-directed behavior.
Oracle success cases—where agents reach near the target but fail to achieve complete success—offer instrumental insight. While all methods eventually approach the correct location, their inability to confidently terminate the episode highlights differences in spatial reasoning and action reliability. These examples expose limitations in termination mechanisms, especially for LLM-based agents operating in a zero-shot regime.
Overall, this qualitative trajectory analysis complements our quantitative evaluation by visualizing policy-specific behaviors in terms of exploration strategy, path geometry, and stop action precision. It reinforces the observed trade-offs between semantic understanding, control consistency, and execution reliability across different agent architectures.


\end{document}